\documentclass[runningheads]{llncs}

\usepackage{eccv}

\usepackage{eccvabbrv}

\usepackage{graphicx}
\usepackage{booktabs}
\usepackage{multirow}
\usepackage{amssymb}
\usepackage{bbding}
\usepackage{enumitem}

\usepackage[accsupp]{axessibility}  % Improves PDF readability for those with disabilities.

\usepackage{hyperref}

\usepackage{orcidlink}

\begin{document}

% ---------------------------------------------------------------
% TODO REVIEW: Replace with your title
\title{Teach CLIP to Develop a Number Sense for Ordinal Regression} 

% TODO REVIEW: If the paper title is too long for the running head, you can set
% an abbreviated paper title here. If not, comment out.
% \titlerunning{Abbreviated paper title}

% TODO FINAL: Replace with your author list. 
% Include the authors' OCRID for the camera-ready version, if at all possible.
\author{Yao Du\inst{1}\orcidlink{0000-0001-7137-8172} \and
Qiang Zhai\inst{1}\orcidlink{0000-0001-5328-675X} \and
Weihang Dai\inst{1}\orcidlink{0000-0002-8619-236X} \and
Xiaomeng Li\inst{1,2}\orcidlink{0000-0003-1105-8083}\thanks{Corresponding Author}}

% TODO FINAL: Replace with an abbreviated list of authors.
\authorrunning{Y.~Du et al.}
% First names are abbreviated in the running head.
% If there are more than two authors, 'et al.' is used.

% TODO FINAL: Replace with your institution list.
\institute{The Hong Kong University of Science and Technology, Hong Kong SAR, China \and
HKUST Shenzhen-Hong Kong Collaborative Innovation Research Institute, Futian, Shenzhen, China}

\maketitle

\begin{abstract}

Ordinal regression is a fundamental problem within the field of computer vision, with customised well-trained models on specific tasks. While pre-trained vision-language models (VLMs) have exhibited impressive performance on various vision tasks, their potential for ordinal regression has received less exploration.
In this study, we first investigate CLIP's potential for ordinal regression, from which we expect the model could generalise to different ordinal regression tasks and scenarios. Unfortunately, vanilla CLIP fails on this task, since current VLMs have a well-documented limitation of encapsulating compositional concepts such as number sense.
% We point out that there are two major reasons for the poor quantitative understanding performance: the presence of insufficient numerical captions during pre-training and the utilisation of an ineffective training objective.
% One is \textbf{\textit{the insufficient number-specific text captions during the pre-training}}, and the other is that original contrastive learning aims to \textbf{\textit{align a whole sentence with an image, which is not sensitive to capture the partial information inside a sentence structure, like the numbers}}.
% Corresponding to the reasons pointed out,
We propose a simple yet effective method called NumCLIP to improve the quantitative understanding of VLMs.
% For the insufficient number captions issue, 
% it is impractical to construct adequate number-specific pairs covering the entire range of values for fine-tuning, considering the rapid increasing of training cost by constructing both positive and negative samples with different numbers. Alternatively, 
We disassemble the exact image to number-specific text matching problem
into coarse classification and fine prediction stages. We discretize and phrase each numerical bin with common language concept to better leverage the available pre-trained alignment in CLIP.
To consider the inherent continuous property of ordinal regression, we propose a novel fine-grained cross-modal ranking-based regularisation loss specifically designed to keep both semantic and ordinal alignment in CLIP's feature space. Experimental results on three general ordinal regression tasks demonstrate the effectiveness of NumCLIP, with 10\% and 3.83\% accuracy improvement on historical image dating and image aesthetics assessment task, respectively. Code is publicly available at \url{https://github.com/xmed-lab/NumCLIP}.
% \begin{figure}[t]
% \begin{center}
%    \includegraphics[width=\linewidth]{figs/hint.png}
% \end{center}
%    \caption{NumCLIP is inspired by the fact that vanilla CLIP has a weak number sense. The zero-shot Mean Absolute Error (MAE) for age estimation is 6.09 and zero-shot accuracy for historical image dating is merely 26.08\%, listed in ``Zero-shot CLIP'' in Table~\ref{table:Result:MORPH} and Table~\ref{table:Result:Historical}. By converting the prompt logic from pure number to common language concept (from quantitative way to qualitative way), the task difficulty can be significantly reduced.} 
% \label{fig:hint}
% \vspace{0.2cm}
% \end{figure}
\end{abstract}
    
\section{Introduction}
\label{sec:intro}

% intro for ordianl regression task
Ordinal regression, also known as ordinal classification, is a machine learning task designed to predict labels with an inherent ordinal order. For example, age estimation involves predicting the age of a facial image, which naturally follows a progression in chronological order~\cite{niu2016ordinal}. Similarly, historical image dating aims to predict the decade of a given historical colored image, which exhibits an inherent ordinal structure~\cite{palermo2012dating}. Ordinal regression is a fundamental problem and has received increasing attention~\cite{li2021learning,deng2021pml,lee2022geometric}.

% intro for current method
Current techniques for ordinal regression can be categorised into three main groups: regression-based, classification-based, and ranking-based methods. Regression based methods involve the direct estimation of a scalar value, typically achieved by minimising the difference through loss functions such as Euclidean loss (e.g., $\ell_1$ or $\ell_2$ loss)~\cite{yi2014age}. Although these methods are straightforward, they usually suffer from subpar performance compared to classification-based methods~\cite{rothe2018deep}.
In early classification-based approaches, cross-entropy loss is commonly used to optimise the network, treating different categories as independent classes~\cite{rothe2015dex}. Recent works have taken into account the ordering relationship among labels. This is achieved through reformulating the single one-hot label as a label distribution~\cite{gao2017deep,pan2018mean} or re-weighting labels based on the continuous relationship~\cite{li2019bridgenet,dai2024semi,dai2021adaptive}.
Ranking-based methods, despite receiving comparatively less attention than the other two categories, provide an alternative approach. 
The underlying concept behind ranking-based methods is that making relative comparisons between samples and anchors should be more manageable than making direct predictions. Therefore, order learning algorithms can be employed in such tasks~\cite{lee2022order,shin2022moving}.

The methods mentioned above primarily focus on learning ordinal concepts within the image domain, which are vulnerable to overfitting. In contrast, recent advancements in pre-trained VLMs, exemplified by CLIP~\cite{radford2021learning}, have shown significant improvements across various downstream tasks~\cite{zhou2022conditional,zhou2023zegclip, gu2021open,lin2022frozen}.
Nonetheless, the potential of CLIP in ordinal regression, a fundamental yet crucial task, remains largely underexplored.
A straightforward approach to applying CLIP for ordinal regression is to treat the numerical index label as the class token and utilise zero/few-shot learning adaptation to make predictions. 
Despite the promising results achieved by applying such a paradigm in CLIP-based common classification/segmentation tasks, which heavily rely on the model's recognition ability, its performance on ordinal regression is notably limited. For instance, in age estimation, the zero-shot MAE is 6.09, and for historical image dating, the zero-shot accuracy is merely 26.08\%; see results of ``Zero-shot CLIP'' in Table~\ref{table:Result:MORPH} and Table~\ref{table:Result:Historical}.
While general prompt learning techniques like CoOp (context optimisation)~\cite{zhou2022conditional} aim to enhance model adaptation to downstream tasks through trainable prompt embeddings, there still exists a significant performance gap when compared to state-of-the-art (SOTA) customised ordinal regression methods such as POE~\cite{li2021learning} and MWR~\cite{shin2022moving}. 

In this study, we attribute the unsatisfactory performance of CLIP-based ordinal regression to two main reasons: the presence of insufficient numerical captions during pre-training and the utilisation of an ineffective training objective. The insufficient numerical captions issue is mainly caused by the fact that captions with specific numbers become rare with the increasing of number magnitude. In other words, when representing larger numbers, it is more frequent and reasonable to use approximate and qualitative descriptions rather than quantitative numbers. The training inefficiency with numbers means that the original contrastive learning for CLIP pre-training aims to align the text embedding of the whole sentence with paired image. The number inside a sentence is only partial information or attribution, which is not explicitly discriminated in current VLM training objective.

To address above challenges, we propose NumCLIP, targeted at developing the number sense of CLIP for ordinal regression. For the numerical caption insufficiency, as mentioned, it is impractical to construct adequate number-specific captions with paired images, leading to the increasing training cost issue. Recalling that qualitative or approximate textual descriptions are frequently occurring captions and probably utilised in the CLIP pre-training (experiment results in Table~\ref{table:few_shot:MORPH} validate this point). Therefore, we discretize the numbers into separate bins and phrase each bin with common language description. Ideally, these linguistic captions should have closer alignment with images, compared with pure numbers. In this way, we can better leverage the available alignment capability to conduct coarse classification using these captions, and then refine the predictions via a coarse-to-fine manner. Meantime, such conversion mitigates the insufficient number caption issue. To regularise the feature representation focusing on discriminating the ordinal concept, we introduce a novel fine-grained cross-modal ranking-based feature regularisation loss. This loss function takes into account the inherent sequential nature of ordinal labels and aims to encourage both semantic and ordinal alignment within CLIP's feature space.

OrdinalCLIP~\cite{li2022ordinalclip} and L2RCLIP~\cite{wang2023learning} also incorporate language priors for ordinal regression task, demonstrating satisfactory performance. However, our method distinguishes itself from the above two methods in two ways. Firstly, OrdinalCLIP~\cite{li2022ordinalclip} and L2RCLIP~\cite{wang2023learning} focus on learning ordering rank prompts from scratch, while our method emphasises the effectiveness of exploring pretrained number-related captions. Secondly, OrdinalCLIP~\cite{li2022ordinalclip} has a weakened semantic alignment, and L2RCLIP~\cite{wang2023learning} proposes an ordinal pairwise loss to address this issue. In contrast, our NumCLIP provides a new perspective by extending mutual information analysis to ordinal regression problem and obtains a better performance.

We summarize the contributions of this paper as follows:
\begin{itemize}
\item
NumCLIP combines the advantages of coarse-to-fine learning, recasting weak image-to-number alignment as strong image-to-language alignment, and fine-grained feature regularisation to teach CLIP to develop a strong number sense and consequently improve the performance for ordinal regression.

\item 
To the best of our knowledge, our work is the first to extend cross-modal contrastive learning for CLIP in ordinal regression task by considering the ordinal relationship among samples and we provide theoretical analysis from mutual information perspective.

\item
Detailed experiments show that NumCLIP outperforms prior SOTA methods on three widely used benchmarks, with 10\% overall accuracy improvement on historical image dating, 3.83\% overall accuracy improvement on image aesthetics assessment, and 1.33 MAE reduction on age estimation under 1-shot setting.

\end{itemize}

\section{Related Work}
\label{sec:related work}
\subsubsection{Ordinal Regression.} The goal of ordinal regression is to learn a rule to map an input image to a rank on an ordinal scale. Regression-based approaches typically employ a Euclidean loss to estimate the precise value, penalising the disparity between predicted values and ground-truth labels~\cite{dai2023semi_}. Yi \emph{et al.}~\cite{yi2014age} propose a multi-scale network to directly estimate ages using $\ell_2$ loss. However, these methods often yield subpar performance as they fail to account for the ordinal relationship among labels.
Classification-based approaches partition the numbers into distinct groups and subsequently treat the estimation of group labels as a classification task. Rothe \emph{et al.}~\cite{rothe2015dex} formulate age estimation as a deep classification task, where the results are further refined using \emph{softmax}-normalised probabilities. This strategy has been shown to outperform direct regression methods.
The aforementioned studies fail to consider the inherent ordering relationship among labels. In contrast, Gao \emph{et al.}~\cite{gao2017deep} address this limitation by modeling the label distribution as a normal distribution centred around the true value and subsequently perform multi-class classification.
Ranking-based approaches, on the other hand, treat the original labels as rank-ordered data and compare the input with multiple reference instances. Shin \emph{et al.}~\cite{shin2022moving} propose a moving window regression algorithm. This algorithm constructs a search window comprising two reference instances and iteratively estimates the relative rank of the input image within the window. 
Unlike previous works that are solely trained within the image domain and lack of generalisation across different tasks and scenarios, our method, NumCLIP, leverages the rich cross-modal image-text knowledge to build a generalised ordinal regression model.
% ------------------------------111111111111111--------------------%
% \vspace{-0.3cm}
\subsubsection{CLIP in Regression.} Recently, there are several attempts to employ CLIP for various regression topics, including depth estimation~\cite{zhang2022can}, crowd/object counting~\cite{liang2023crowdclip,paiss2023teaching}, and ordinal regression~\cite{li2022ordinalclip}.
Depth estimation is a dense prediction task to infer the depth for each pixel, and normally the physical environment variance, depth changing rate with distance, and strategies for minimising computational complexity are considered to ensure the satisfactory performance~\cite{fu2018deep,bhat2021adabins}. Similar situations exist for counting tasks, such as considering the object density variation in different areas~\cite{liu2019context}. 
DepthCLIP~\cite{zhang2022can} investigates CLIP-based monocular depth estimation in a zero-shot manner, while CrowdCLIP~\cite{liang2023crowdclip} explores CLIP's potential on crowd counting. 
CountingCLIP~\cite{paiss2023teaching} proposes a counting-contrastive loss to teach CLIP to count to ten.
In general, these two tasks are well-defined domain-specific tasks. 
By contrast, ordinal regression is a fundamental task that involves estimating category labels with an inherent ordering relationship. This task is applicable to various domains, such as age estimation, historical image dating, and image aesthetics assessment. In this study, our focus is specifically on the task of ordinal regression.
OrdinalCLIP~\cite{li2022ordinalclip} and L2RCLIP~\cite{wang2023learning} focus on modelling the ordering property of rank prompts while overlooking the available pre-trained number-related knowledge.
In contrast, our approach, NumCLIP, outperforms OrdinalCLIP~\cite{li2022ordinalclip} and L2RCLIP~\cite{wang2023learning} with a large margin by effectively utilised pre-trained alignment and ordinality-targeted training objective.
% ------------------------------111111111111111--------------------%
% \vspace{-0.3cm}
\subsubsection{Compositionality of VLMs.}\label{sec: composition}
Despite the impressive achievements, more recent works point out that current VLMs like CLIP have a weak understanding of fine-grained concepts like relational, compositional, and contextual reasoning~\cite{radford2021learning,paiss2023teaching,kamath2023text,xu2023challenges,paiss2022no}. Radford \emph{et al.}~\cite{radford2021learning} state that CLIP is poor on fine-grained classification tasks and struggles with more systematic and abstract concepts such as counting the exact number of objects. Paiss \emph{et al.}~\cite{paiss2022no, paiss2023teaching} demonstrate that CLIP only partially captures the information of input text, and is less responsive to prepositions, numbers, and adjectives. Kamath \emph{et al.}~\cite{kamath2023text} find that the text encoder of CLIP falls short on attribute-object association, negation, object relationship and counting. Yuksekgonul \emph{et al.}~\cite{yuksekgonul2022and} find that current VLMs have a poor relationship understanding, blunder when linking objects with attributes, and demonstrate a severe lack of order sensitivity.
While recent works have primarily focused on addressing compositional limitations, effective improvements for CLIP-based ordinal regression have not been widely investigated.
\section{Method}
\subsection{Problem Statement}
A popular baseline for ordinal regression is to convert it as a classification task by discretizing the labels as different bins and treating each bin as an independent class. After problem reformulation, typical multi-class classification losses like cross-entropy loss are adopted for model training and the final prediction values could be obtained by either choosing the class index with highest probability or linearly multiplying all class probabilities with corresponding index values. Extra regularisation might be considered by utilising the ordinal relationship among labels.
Mathematically, let $x_i$ denote the $i$ input instance with corresponding discretized label $y_i$, ordinal regression aims to recover $y_i$ by encoding the input image into feature $z_i = \Phi(x_i)$ and using a classifier $f(\cdot)$ to compute class probability $p_i$. The model is optimised with cross-entropy loss. The adaptation of VLMs like CLIP to boost ordinal regression could be achieved via image-text matching in a coarse-to-fine manner. Specifically, image feature $z_i$ could be extracted using a pre-trained CLIP image encoder by $z_i = image(x_i)$. For a given ordinal regression task, task-related text description $r_i$ is constructed based on linguistic mapping. Such templates are converted into fixed-size tokens with $tokenizer(\cdot)$ and mapped into text embeddings $w_i$ with CLIP text encoder. The process can be formulated as: $w_i = text(tokenizer(r_i))$. These text embeddings are regarded as
classifier weights in a typical classification task to make coarse classification. After that, a lightweight MLP regressor may concatenate to refine the coarse estimation into fine-grained predictions.

\begin{figure}[t]
\begin{center}
   \includegraphics[width=1.0\linewidth]{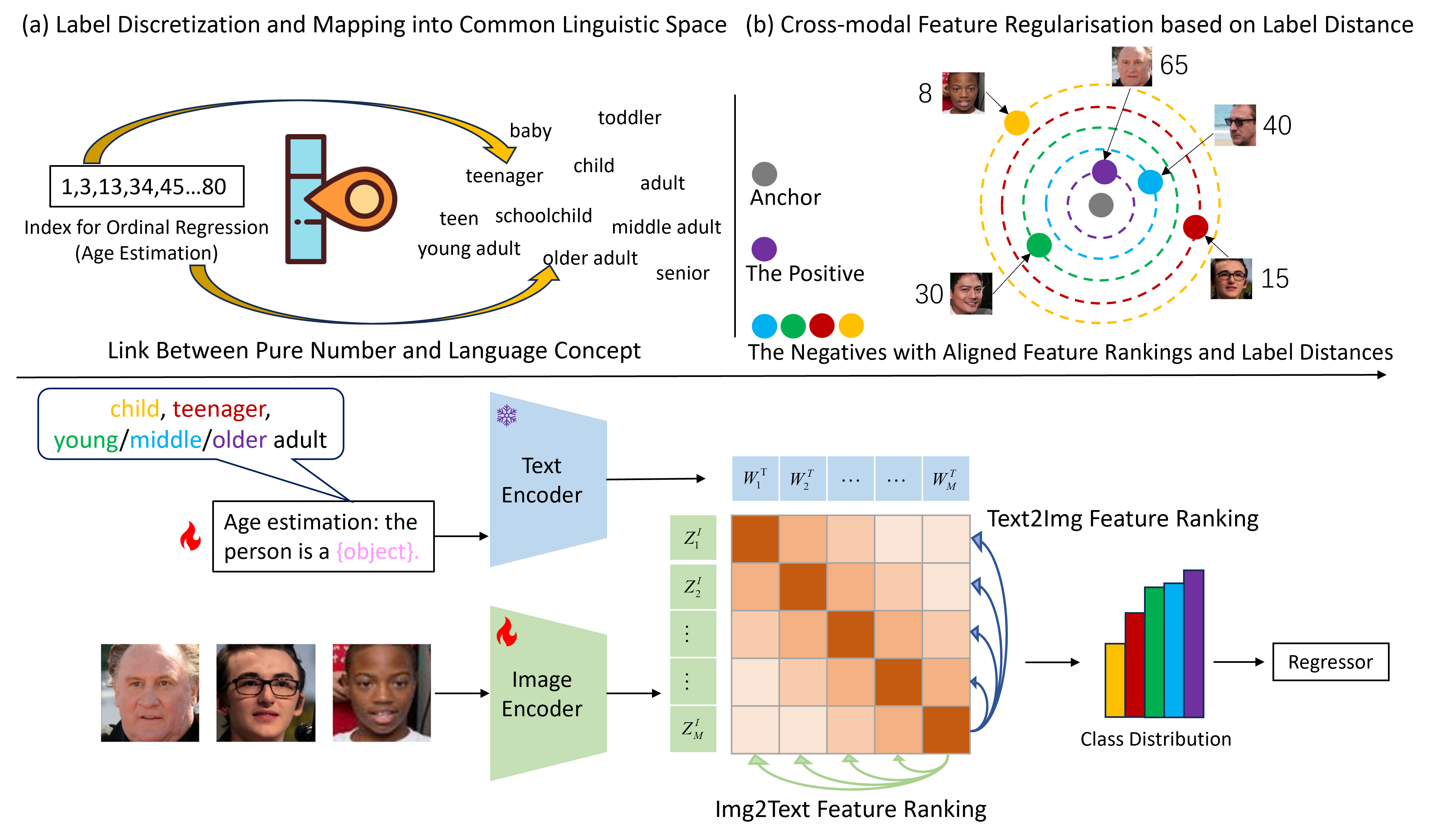}
\end{center}
   \caption{The framework of NumCLIP, aiming to teach CLIP to develop a strong number sense for ordinal regression. Replacing pure numbers as common language descriptions allow better utilising the pre-training knowledge, and cross-modal ranking-based feature regularisation ensures both semantic and ordinal alignment.
   }
\label{fig:framework}
% \vspace{-0.2cm}
\end{figure}

\subsection{Coarse-to-Fine Ordinal Regression Paradigm}
Overall, we design a coarse-to-fine CLIP-based ordinal regression paradigm, shown in Figure~\ref{fig:framework}. The motivation for this is based on the fact that learning from a staged classification process is more effective and easier than directly learning from multiple precise values, especially in the imperfect data scenario. In the coarse stage, we first discretize the continuous numbers into different bins, with each bin linguistically mapped with a common textual concept/description, as shown in Part(a) of Figure~\ref{fig:framework}. From that we elegantly convert a dense regression task into a simple and coarse classification problem, which not only smoothly mitigates the insufficient number caption issue, but also effectively utilises/recalls the pre-trained/available concept alignment learned by CLIP. The output of coarse stage is the logits between image and text feature similarities.
There are two levels of fine stage. The first is the cross-modality ranking-based feature regularisation loss to refine the feature alignment with an ordinal structure, as shown in Part(b) of Figure~\ref{fig:framework}. The other is the concatenated light regressor in case that the exact numerical prediction is expected, where the results could be derived from the classification distribution and the regressor is learnable.
The final prediction is calculated as: 
\begin{equation}
    y^{*}= \sum\limits_{i=1:k} p_i * \frac{b_i}{1+{\delta}_i}
\end{equation}
where $k$ is the number of classes, $p_i$ is the class probability, $b_i$ is the centre of $i_{th}$ mapped numerical group, and ${\delta}_i$ is the estimated shift from the regressor to make the bin interval learnable.
These two stages are trained end-to-end.

\subsection{Mapping from Number to Linguistic Concept}
\label{sec: free lunch}

\subsubsection{Insufficient Number Captions during Pre-training.} 
\begin{figure}[t]
\centering
\scalebox{0.6}{
% \begin{center}
   \includegraphics[width=\linewidth]{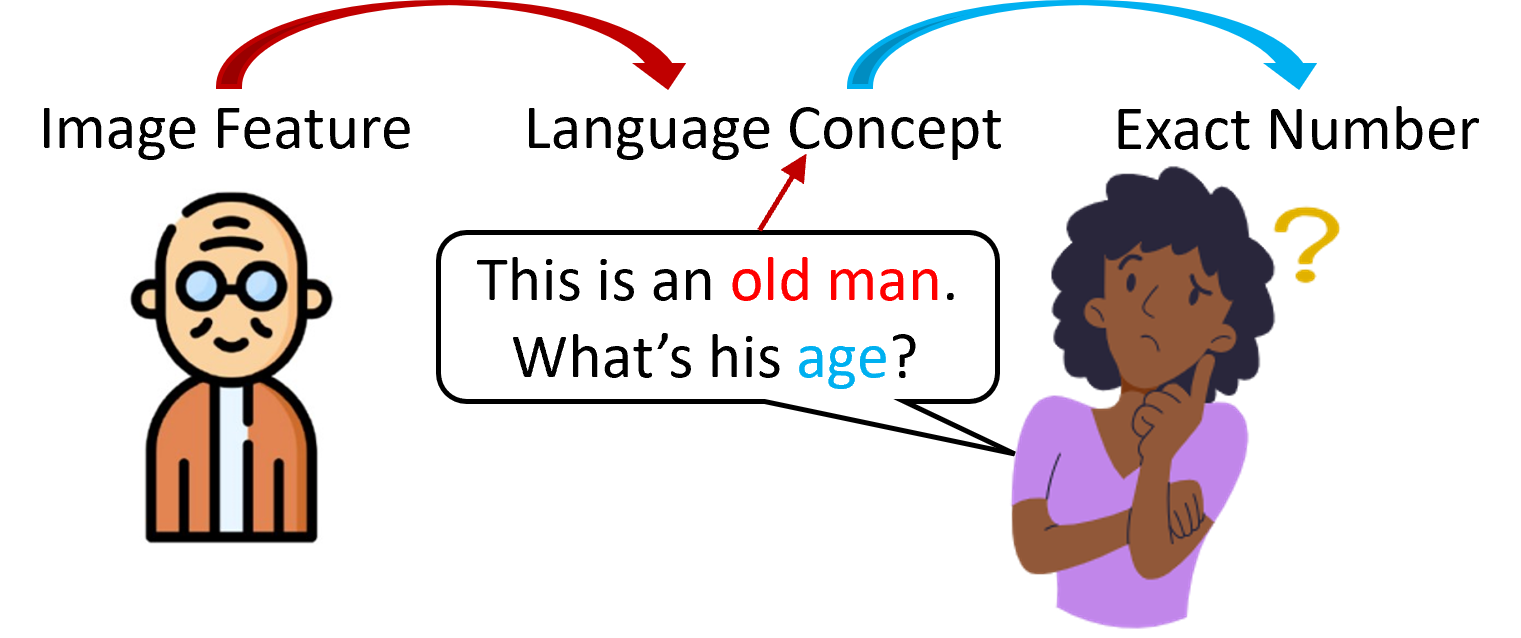}
% \end{center} 
}
   \caption{Mimic human numerical cognition: mapping an image feature to a language concept first, and then reasoning the number.
   }
   
\label{fig:img-lang-num}
% \vspace{-0.2cm}
\end{figure}

As stated in Section \ref{sec: composition}, current VLMs fall short on compositional or fine-grained concepts, 
struggling with number-related tasks~\cite{paiss2023teaching,paiss2022no}.
One major reason is that there are insufficient captions for exact number descriptions with paired images from pre-trained dataset, especially when the number magnitude is large, leading to unsatisfactory alignment between images and numbers. 
% From another perspective, in classification or segmentation, the object class name is a constant language concept, and will not change with different tasks, datasets, or scenarios. In contrast, \textbf{\textit{the ordinal index/number itself represents a relative comparison}} and the concept will be different if the object, measuring metric, or other related factors vary, even though the number itself is still the same. Therefore, it is difficult to train a well-aligned model to match image features with a changing concept. 
% \textbf{\textit{It seems difficult, if not impossible, to replicate the remarkable success of CLIP-based classification/segmentation for ordinal regression task in a same/similar processing manner}}.
Let's see how human beings solve this task from a numerical cognition perspective~\cite{lipton2003origins,kaufman1949discrimination}. There are two number representation systems in human cognition: an accurate and confident system to discriminate small numerosities (i.e., 1-4 items), referred to as subitizing and mainly based on visual information to directly link image feature with a number concept, and an approximate system to represent larger quantities based on intuitive reasoning, known as number sense, to map the image feature with a language concept first, and then derive the specific number if necessary. Such observations support that it is not reasonable to directly treat numbers as class tokens and simply regard the ordinal regression problem as an image-text matching task. In contrast, we wish to teach CLIP with a strong number sense, as shown in Figure~\ref{fig:img-lang-num}.

% --------------------------------1111111----------------------
% \vspace{-0.3cm}
\subsubsection{Recast the Tough Ordinal Regression as Simple Classification.}
Following the number sense concept above~\cite{lipton2003origins,kaufman1949discrimination}, we cast numerical indexes into common language space based on the linguistic common knowledge from specific tasks, instead of rigidly learning image-to-number alignment from scratch, as shown in Figure~\ref{fig:img-lang-num}. 
The reasons for mapping numbers into linguistic semantic concepts are three-folds: First is the potential large quantity of numerical range, leading to the redundancy and infeasibility of treating each number as a class token for training. Second is the concept-changing nature of numbers with respect to different tasks, measuring metrics, etc. In contrast, language description has a stronger resistance to these variations, as we discussed above. Third is the free ``take-away'' from pre-trained VLMs. Taking age estimation as an example, we can use ``older adult'' or ``teenager'' to describe a person instead of specific ages, shown in Part(a) of Figure~\ref{fig:framework}. It is reasonable to assume that such number-related quantifiable descriptions appear more frequently in the pre-trained dataset, and thus are desired to make stronger responses/alignments than single numbers (see results in Table~\ref{table:ablation}).

Specifically, we initialise the language-to-number mapping relationship by querying LLMs with a prompt as: ``how to describe people with different ages?", from which we obtain descriptions for different age ranges. For each data in training set, we match corresponding language concept according to the age range that the ground-truth label falls in. Such initialised prompts are learnable by converting into a set of learnable vectors like CoOp~\cite{zhou2022conditional} with end-to-end training to ensure correctness. These prompts are fixed in testing to avoid randomness.
This section provides a new perspective for VLM numerical reasoning by mimicking human cognition, rather than focuses designing dedicated prompts like general CLIP adaptation methods.
% --------------------------------1111111----------------------
% \vspace{-0.3cm}
\subsection{Cross-Modal Ranking-based Feature Regularisation}
\label{sec: cross-modal}

\subsubsection{Overview.}
The transition from numbers to linguistic concepts addresses the problem of insufficient numerical training captions, while the insensitivity of fine-grained descriptions in the original contrastive learning (stated in Section \ref{sec: composition}) will be solved by our proposed cross-modal ranking-based feature regularizer to encourage both semantic and ordinal alignment in CLIP's feature space.

\begin{figure}[t]
\centering
\scalebox{0.5}{
% \begin{center}
   \includegraphics[width=\linewidth]{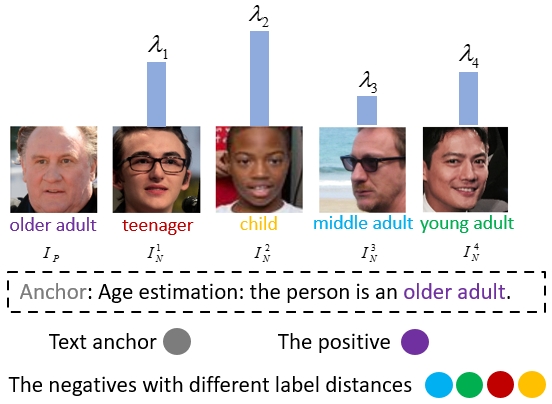}
% \end{center}
}
   \caption{Fine-grained cross-modal ranking-based feature regularisation. The cross-modal negative samples are pushed away with ordinal label distance alignment.
   }
   
\label{fig:rank_loss}
% \vspace{-0.5cm}
\end{figure}

Specifically, current vision-language pre-training (VLP) is typically conducted with cross-modal contrastive learning, e.g., the InfoNCE loss~\cite{oord2018representation}. Taking the text-to-image contrastive loss as an example, given an image-text pair $(I,T)$ with $T$ being the anchor and $I_P$ being the positive image sample, all other images in a mini-batch will be regarded as negative samples and therefore be pushed away from the anchor. This training objective could be problematic for ordinal regression task since the inherent ordering relationship among adjacent samples is ignored. As shown in Figure~\ref{fig:rank_loss}, a text can be semantically paired with multiple images with different errors and the subscript $N$ indicates the negative sample. Similar to classification, the prediction error of misclassifying ``older adult'' as ``middle adult'' should be lower than that of misclassifying as ``teenager'' since ``middle adult'' is closer to ``order adult''. Therefore, indiscriminating the label distances between an image/text anchor with its all negative texts/images will inevitably hinder the learning effect, leading to suboptimal cross-modal representation for ordinal regression. 

\subsubsection{Analysis from Mutual Information Perspective.}
The InfoNCE loss function, commonly used in contrastive learning, can be interpreted as a lower bound approximation of mutual information (MI) between anchors and positive samples~\cite{oord2018representation}. 
Formally, with a batch of M semantically aligned image-text pairs $\{\left(I_i, T_i\right)\}_{i=1:M}$ and $\ell_2$ normalised embeddings $z^{i=1:M}$ and $w^{i=1:M}$ of each image and text in the batch, the similarity function $f\left( z_i, w_i\right)$ can be utilised to model the density ratio to preserve the MI between $z_i$ and $w_i$. Rewriting the $f\left( z_i, w_i\right)$ to $ \frac{\mathop{P}\left( w_i | z_i \right)}{\mathop{P}\left( w_i\right)} $, the well-known lower bound of MI between $w_i$ and $z_i$ could be derived:
\begin{normalsize}
\begin{equation}
    \mathop{I}(w_i, z_i) \geq  log\left(M\right) - \mathcal{L}_{InfoNCE}^z
   \label{eq2}
\end{equation}
\end{normalsize}
where $\mathop{I}(w_i, z_i)$ is the MI between $w_i$ and $z_i$.
The detailed proof of the derivation could be found in~\cite{oord2018representation}.

% \begin{figure}[tb]
%   \centering
%   \begin{subfigure}{0.48\linewidth}
%     \fbox{\rule{0pt}{0.5in} \rule{.9\linewidth}{0pt}}
%     \caption{An example of a subfigure}
%     \label{fig:short-a}
%   \end{subfigure}
%   \hfill
%   \begin{subfigure}{0.48\linewidth}
%     % \fbox{\rule{0pt}{0.5in} \rule{.9\linewidth}{0pt}}
%     \includegraphics[width=\linewidth]{figs/ranking_loss.png}
%     \caption{Fine-grained cross-modal ranking-based feature regularisation. The cross-modal negative samples are pushed away with ordinal label distance alignment.}
%     \label{fig:short-b}
%   \end{subfigure}
%   \label{fig:short}
% \end{figure}

% \begin{figure}[htbp]
% \centering
% \begin{minipage}[t]{0.48\textwidth}
% \centering
% \includegraphics[width=6cm]{figs/img_lang_num.png}
% \caption{Mimic human numerical cognition: mapping an image feature to a language concept first, and then reasoning the number.}
% \end{minipage}
% \begin{minipage}[t]{0.48\textwidth}
% \centering
% \includegraphics[width=6cm]{figs/ranking_loss.png}
% \caption{Fine-grained cross-modal ranking-based feature regularisation. The cross-modal negative samples are pushed away with ordinal label distance alignment.}
% \end{minipage}
% \end{figure}

\subsubsection{Extend MI Analysis to Ordinal Regression.}
The above derivation implicitly assumes the independence of $w_j$ (the negative sample) and $z_i$, which is reasonable when encountering common classification/discrimination tasks. As a result, the expectation of density ratio $ \frac{\mathop{P}\left( w_j | z_i \right)}{\mathop{P}\left( w_j\right)} $  is equal to 1 and can be disregarded. However, in scenarios where negative samples hold different label distance relationships with the anchor, seen the example in Part(b) of Figure~\ref{fig:framework} and Figure~\ref{fig:rank_loss}, the false negatives are non-negligible, $w_j$  and $z_i$ may not be independent. Therefore, we revisit this derivation to establish a more general formulation:
\begin{normalsize}
\begin{equation}
    \mathop{I}(w_i, z_i) - \mathop{E}\limits_{w_j}\mathop{I}(w_j, z_i) \geq  log\left(M\right) -  \mathcal{L}_{InfoNCE}^z 
    \label{eq3}
\end{equation}
\end{normalsize}
Equation~\ref{eq3} provides a more general lower bound form that the InfoNCE loss optimises. The left side of this equation encompasses two terms: the MI between an anchor and the positive, and MI expectation between an anchor and all negatives. Equation~\ref{eq3} demonstrates that optimising InfoNCE is equivalent to maximising the lower bound of the difference between these two quantities. By combining Equation \ref{eq2} and \ref{eq3}, it can be observed that besides maximising MI between an anchor and the positive (referred to as \textit{MI-P}), InfoNCE loss also aims to minimise the MI expectation between an anchor and negatives (referred to as \textit{MI-N}), ideally approaching zero. However, it should be noted that false negatives may exhibit semantic similarity with the anchor, like the neighbouring label relationship in ordinal regression task, and excessively minimising~\textit{MI-N} could result in a less structural cross-modal representation space. 
To address this issue, a prior target estimation of \textit{MI-N} should be determined. Regarding this, we leverage cross-modal label distance ranking to approximate MI between an image and text. Another challenge lies in integrating this prior estimation into the optimisation process. Based on the derivation of Equation \ref{eq3}, it can be proved that assigning a positive weight $\lambda_{i,j}$ to each $f\left( z_i, w_j\right)$ can effectively control ~\textit{MI-N} to a desirable positive value, subject to the following two conditions:
\begin{itemize}
    \item Condition 1. The covariance between $\lambda_{i,j}$ and $\frac{\mathop{P}\left( w_j | z_i \right)}{\mathop{P}\left( w_j\right)}$ is negative.
    \item Condition 2. The expectation of $\lambda_{i,j}$ among all negative samples is equal to 1. 
\end{itemize}

\subsubsection{Practical Implementation for Ordinal Regression.}
Following the above derivation, we introduce a weight parameter $\lambda_{i,j}$ for each negative sample, which is derived based on cross-modal label distance ranking. We propose Fine-grained Cross-modal Ranking-based Contrastive loss (FCRC) to keep the negative samples from being improperly minimised and boost a more semantically meaningful representation space aligned with ordinal property:
\begin{small}
\begin{align}
   \mathcal{L}_{FCRC}^z  = -\sum\limits_{i=1:M} \frac{1}{M}\mathop{log} 
                                \left[
                                \frac{
                                    f\left( z^i, w^i\right)
                                }
                                {
                                    f\left( z^i, w^i\right)+ \sum\limits_{j \neq i} 
                                    \lambda_{i,j}^z * 
                                    f\left( z^i, w^j\right)
                                 }
                                \right]
% \label{eq1}
\end{align}
\end{small}
where $f\left( z^i, w^j\right) = exp \left( cos\left( z_i, w_j\right)/\tau\right)$ and $\lambda_{i,j}^z$ indicates the regularisation weight of $j_{th}$ negative text sample with respect to $i_{th}$ image anchor in the ordinal regression framework. The FCRC loss of text-to-image can be written with a similar format.
% \begin{small}
% \begin{align}
%     \mathcal{L}_{FCRC}^w =-\sum\limits_{i=1:M} \frac{1}{M}\mathop{log} 
%                                 \left[
%                                 \frac{
%                                     f\left( w^i, z^i\right)
%                                 }
%                                 {
%                                     f\left( w^i, z^i\right)+ \sum\limits_{j \neq i} 
%                                     \lambda_{i,j}^w * 
%                                     f\left( w^i, z^j\right)
%                                  }
%                                 \right]
% % \label{eq2}
% \end{align}
% \end{small}
% where $f\left( w^i, z^j\right) = exp \left( cos\left( w_i, z_j\right)/\tau\right)$ and $\lambda_{i,j}^w$ indicates the contrastive weight of $j_{th}$ negative image sample with respect to $i_{th}$ text anchor in the ordinal training framework.

We adopt an efficient and simplified way to calculate the regularisation weight parameter $\lambda_{i,j}$ of negative samples. Mathematically, the weight parameter $\lambda_{i,j}$ can be derived as follows:
\begin{equation}
    \lambda_{i,j} = Norm(\beta * {d_{i,j}})
\end{equation}
\begin{equation}
    {d_{i,j}} = \left|{y_{i}}-{y_{j}}\right|
\end{equation}
where $\beta$ is a scaling factor, $d_{i,j}$ is the absolute label distance between the anchor and its negative sample, and the \emph{Norm} function is to ensure the average of all weighted terms is equal to 1.

\subsection{Overall Training Objective}
Given that NumCLIP follows a coarse-to-fine ordinal regression paradigm, the overall training objective is a revised Fine-grained Cross-modal Ranking-based Contrastive loss (FCRC) to refine the coarse estimation with the ordinal property constraint, and a regular Euclidean regression loss like MAE to achieve the fine-grained prediction. The weight between FCRC and regression loss is set to 1 and the model can be trained in an end-to-end way.

\section{Experiment}
\subsection{Datasets and Experiment Settings}
\subsubsection{Datasets.}
We conduct detailed experiments on three different and widely-adopted ordinal regression benchmarks, including age estimation, historical image dating and image aesthetics assessment.
\begin{itemize}[leftmargin=*]
    \item \emph{Age Estimation}: The widely used MORPH II~\cite{ricanek2006morph} dataset is adopted, which contains 55,134 portraits from 13,618 individuals. Each portrait image is labeled with an age value from 16 to 77. Following popular evaluation protocols~\cite{rothe2018deep,li2019bridgenet}, only 5,492 images of Caucasian descent are used to remove cross-race interference. 
    \item \emph{Historical Image Dating}: The historical image dating dataset~\cite{palermo2012dating} is a benchmark for predicting the decade of given historical colored image. There are five decade categories from the 1930s to 1970s, where each category contains 265 images. We follow the general ordinal regression settings~\cite{liu2018constrained,liu2019probabilistic}.
    \item \emph{Image Aesthetics Assessment}: The image aesthetic dataset~\cite{schifanella2015image} contains 13,929 available Flickr photos of four categories, including nature, animal, urban, and people. Each image is judged by at least five examiners, and five absolute rating scores are used to evaluate the aesthetics quality: ``unacceptable'', ``flawed'', ``ordinary'', ``professional'', and ``exceptional''. The ground-truth label of each image is set to be the median among its all gradings. We follow the general ordinal regression setting~\cite{wang2023learning}.
\end{itemize}

\subsubsection{Experiment Setting.}
We adopt the ViT-B/16 image and text encoder of CLIP~\cite{radford2021learning} as the model backbone for all experiments. Following OrdinalCLIP~\cite{li2022ordinalclip}, all training data are first resized into 256 × 256 and then randomly cropped into 224 × 224. We take random horizontal flipping as additional data augmentation. The batch size is 32. We train the model for 100 epochs with Adam~\cite{kingma2014adam}. The text encoder is frozen to keep the pre-trained language semantics intact, only the prompt, image encoder and regressor are trained with a small learning rate of 1e-5. All experiments are conducted on a single NVIDIA 3090 GPU.

\subsection{Results under Fully Fine-tuning Setting}
\label{sec: main results}

\textbf{Inferring Age from Images.}
The upper part of Table~\ref{table:Result:MORPH} presents the results of existing SOTA methods for ordinal regression.
It becomes apparent that compared to these customised approaches, the zero-shot CLIP only achieves an MAE of 6.09, indicating a limited understanding of numerical or ordinal concept. CoOp~\cite{zhou2022conditional}, which updates the number prompt and achieves an MAE of 2.39, highlights the importance of model fine-tuning.
OrdinalCLIP~\cite{li2022ordinalclip} explicitly models the ranking property of input text embeddings using linear interpolation, resulting in an MAE of 2.32. L2RCLIP~\cite{wang2023learning} performs token-level attention to enhance the ordering relation of the original rank prompts and achieves an MAE of 2.13. Notably, NumCLIP achieves a superior MAE of 2.08, surpassing CLIP-based methods and remaining competitive with well-designed SOTA ordinal regression techniques. 
%Table~\ref{table:Result:MORPH}, which verifies the effectiveness of our method.

  \begin{table*}[tb]
    \begin{minipage}[t]{0.455\textwidth}
    \centering\small
        \caption{Results of age estimation on MORPH II.}
        \label{table:Result:MORPH}
          \renewcommand\tabcolsep{3pt}
          \scalebox{0.8}{
            \begin{tabular}{lcc}
            \toprule
            Methods  & MAE & {{\color[HTML]{CB0000} $\Delta$}} \\
            \midrule
            % AGEn~\cite{tan2017efficient}& 2.52 & 2018 \\
            % BridgeNet~\cite{li2019bridgenet} & 2.38 & 2019 \\ 
            AVDL~\cite{wen2020adaptive} & 2.37 & {\color[HTML]{CB0000} 0.29} \\
            DRC-ORID~\cite{lee2020deep} & 2.26 & {\color[HTML]{CB0000} 0.18} \\
            POE~\cite{li2021learning} & 2.35 & {\color[HTML]{CB0000} 0.27} \\
            PML~\cite{deng2021pml} & 2.31 &  {\color[HTML]{CB0000} 0.23} \\
            MWR~\cite{shin2022moving} & 2.13 &  {\color[HTML]{CB0000} 0.05} \\
            \hline
            Zero-shot CLIP~\cite{radford2021learning} & 6.09 & {\color[HTML]{CB0000} 4.01} \\
            CoOp~\cite{zhou2022conditional} & 2.39 & {\color[HTML]{CB0000} 0.31} \\
            OrdinalCLIP~\cite{li2022ordinalclip} & 2.32 & {\color[HTML]{CB0000} 0.24} \\
            L2RCLIP~\cite{wang2023learning} & 2.13 & {\color[HTML]{CB0000} 0.05} \\
            \textbf{NumCLIP (Ours)} & \textbf{2.08} & - \\
            \bottomrule
            \end{tabular}}
    \end{minipage}
  \hfill
    \begin{minipage}[t]{0.495\textwidth}
    \centering\small
        \caption{Results on Historical Image Dating.}
        \label{table:Result:Historical}
          \renewcommand\tabcolsep{3pt}
          \scalebox{0.8}{
            \begin{tabular}{lcc}
            \toprule
            Methods  & Accuracy (\%) & MAE \\
            \midrule
            CNNPOR~\cite{liu2018constrained}  & 50.12 $\pm$ 2.65 & 0.82 $\pm$ 0.05\\
            % GP-DNNOR~\cite{liu2019probabilistic}  & 46.60 $\pm$ 2.98 & 0.76 $\pm$ 0.05\\
            POE~\cite{li2021learning} & 54.68 $\pm$ 3.21 & 0.67 $\pm$ 0.04   \\
            MWR~\cite{shin2022moving}  & 57.8 & 0.58    \\
            GOL~\cite{lee2022geometric}  & 56.2 & 0.55 \\
            Ord2Seq~\cite{wang2023ord2seq} & 59.5 & 0.53 \\
            \hline
            Zero-shot CLIP~\cite{radford2021learning} & 26.08 $\pm$ 0.56 & 1.48 $\pm$ 0.03 \\
            CoOp~\cite{zhou2022conditional} & 51.90  $\pm$ 2.60 & 0.76 $\pm$ 0.06 \\
            OrdinalCLIP~\cite{li2022ordinalclip} & 56.44 $\pm$ 1.66 & 0.67 $\pm$ 0.03  \\
            L2RCLIP~\cite{wang2023learning} & 67.22 $\pm$ 1.59 & 0.43 $\pm$ 0.03  \\
            \textbf{NumCLIP (Ours)} & \textbf{69.61 $\pm$  2.02} & \textbf{0.35  $\pm$ 0.03} \\
            \bottomrule
            \end{tabular}}
    \end{minipage}
  \end{table*}

\begin{table}[t]
    \centering
    \caption{
Quantitative results on the Image Aesthetic dataset. Both accuracy and MAE are reported.
}
    \scalebox{0.85}{
    \begin{tabular*}{13.8cm}{l|ccccc|ccccc}
        \toprule
         \multirow{2}*{Methods} &
        \multicolumn{5}{c}{Accuracy (\%) -- higher is better} 
        &  \multicolumn{5}{c}{MAE -- lower is better}
        \\
        \cmidrule(r){2-6} \cmidrule(r){7-11}  & Nature & Animal & Urban & People & Overall 
        & Nature & Animal & Urban & People & Overall
        \\
        \midrule
        CNNPOR   \cite{liu2018constrained} & 71.86 & 69.32 & 69.09 & 69.94 & 70.05 & 0.294 & 0.322 & 0.325 & 0.321 & 0.316  \\
        SORD \cite{diaz2019soft} & 73.59 & 70.29 & 73.25 & 70.59 & 72.03 & 0.271 & 0.308 & 0.276 & 0.309 & 0.290 \\
        POE~\cite{li2021learning} & 73.62 & 71.14 & 72.78 & 72.22 & 72.44 & 0.273 & 0.299 & 0.281 & 0.293 & 0.287 \\
        GOL~\cite{lee2022geometric} &73.8 & 72.4 & 74.2 & 69.6 & 72.7 & 0.27 & 0.28 & 0.26 & 0.31 & 0.28 \\
        \midrule
        Zero-shot CLIP~\cite{radford2021learning} & 65.24 & 45.67 & 58.78 & 53.06 & 55.68 & 0.461 & 0.557 & 0.468 & 0.524 & 0.502 \\
        CoOp~\cite{zhou2022conditional} & 72.74 & 71.46 & 72.14 & 69.34 & 71.42 & 0.285 & 0.298 & 0.294 & 0.330 & 0.302  \\
        OrdinalCLIP~\cite{li2022ordinalclip} & 73.65 & 72.85 & 73.20 & 72.50 & 73.05 & 0.273 & 0.279 & 0.277 & 0.291 & 0.280  \\
        L2RCLIP~\cite{wang2023learning} &  73.51  &  \textbf{75.26}  &  77.76  &  \textbf{78.69}  &  76.07    &  0.267  &  0.253  &  0.216  &  0.246  &  0.245     \\
        \textbf{NumCLIP (Ours)} & \textbf{75.20} & 75.24 & \textbf{79.49} & 76.17 & \textbf{76.53} & \textbf{0.249} & \textbf{0.250}	& \textbf{0.208} & \textbf{0.238} &	\textbf{0.236} \\
        \bottomrule
    \end{tabular*}}
% \vspace{-0.5cm}
\label{table:SOTA:aes}
\end{table}

\textbf{Inferring Decade from Images.}
The results presented in Table~\ref{table:Result:Historical} indicate that zero-shot CLIP still exhibits a poor performance due to its limited understanding of numbers. In contrast, CoOp~\cite{zhou2022conditional} and OrdinalCLIP~\cite{li2022ordinalclip} significantly improve the overall performance compared to zero-shot CLIP. L2RCLIP~\cite{wang2023learning} introduces token-level attention for prompt tuning and ordinal pairwise loss, achieving an accuracy of 67.22\% and an MAE of 0.43.
Notably, NumCLIP achieves the highest accuracy of 69.61\% and lowest MAE of 0.35, surpassing all other methods by a significant margin.

\begin{table}[thbp]
\caption{The MAE results under few-shot setting on MORPH II.}
\label{table:few_shot:MORPH}
\centering
\scalebox{0.9}{
\begin{tabular}{l|cccccccc}
\toprule
\# Shots & 1 & 2 & 4 & 8 & 16 & 32 & 64 & all \\
\midrule
CoOp~\cite{zhou2022conditional} & 5.09  & 4.50  & 3.81  & 3.57  & 3.23  & 2.87  & 2.61 & 2.39\\ 
OrdinalCLIP~\cite{li2022ordinalclip} & 4.94  & 4.36  & 3.55  & 3.31  & 3.07  & 2.76  & 2.57 & 2.32 \\
L2RCLIP~\cite{wang2023learning} & 4.54 & 3.92 & 3.40 & 3.28 & 2.81 & \textbf{2.55} & 2.38 & 2.13 \\
\textbf{NumCLIP (Ours)} & \textbf{3.61} & \textbf{3.17} & \textbf{3.20} & \textbf{2.96} & \textbf{2.79} & 2.60 & \textbf{2.38} & \textbf{2.11} \\
\bottomrule
\end{tabular}}
% \vspace{-0.4cm}
\end{table}

\textbf{Inferring Aesthetics Grading from Images.} Table~\ref{table:SOTA:aes} presents the results of image aesthetics grading. Zero-shot CLIP performs poorly, struggling to differentiate between ordinal concepts. CoOp~\cite{zhou2022conditional}, OrdinalCLIP~\cite{li2022ordinalclip} and L2RCLIP~\cite{wang2023learning} exhibit comparable performance to previous best-performing methods. Again, NumCLIP outperforms all other methods, achieving an impressive overall accuracy of 76.53\% and an MAE of 0.236. The results in individual categories also demonstrate satisfactory performance.

\subsection{Results under Few-shot Setting}

Following OrdinalCLIP~\cite{li2022ordinalclip}, few-shot setting is conducted on the MORPH II~\cite{ricanek2006morph} dataset to further validate the model generalisation performance. 
The results are presented in Table~\ref{table:few_shot:MORPH}. It is evident that NumCLIP consistently outperforms other methods by a significant margin, particularly in the 1-shot and 2-shot settings. By formulating the problem as a coarse-to-fine paradigm with consideration of incorporating language priors and aligning semantic features in an ordinal manner, NumCLIP achieves an impressive MAE of 3.61, compared to 5.09 of CoOp~\cite{zhou2022conditional}, 4.94 of OrdinalCLIP~\cite{li2022ordinalclip} and 4.54 of L2RCLIP~\cite{wang2023learning} under the 1-shot setting. Similar performance gains can be observed across other shot settings, highlighting the effectiveness of NumCLIP. This impressive performance demonstrates the label efficiency of NumCLIP, which is crucial for scenarios with limited training data.

\subsection{Ablation Study}
The success of NumCLIP is mainly contributed to three important components, namely coarse-to-fine ordinal regression paradigm, converting pure numbers into common linguistic concepts and fine-grained cross-modal ranking-based regularizer. Since the proposed coarse-to-fine paradigm is fundamental yet distinct with previous SOTA methods with end-to-end classification, we select it as our baseline to validate the effectiveness of the dedicatedly designed two components, where the class prompts are randomly initialised and the training objective is original contrastive learning loss with regression loss. It is worth mentioning that even without any additional modules, our baseline obtains an MAE of 2.25, which is already higher than 2.32 
of OrdinalCLIP~\cite{li2022ordinalclip} listed in Table~\ref{table:Result:MORPH}, under the same setting. This impressive result indicates the effectiveness of the coarse-to-fine paradigm to reduce the learning difficulty, especially for tasks with multiple classes, such as age estimation.

\begin{figure}[thbp]
  \centering
  \subfloat[Vanilla CLIP]{\includegraphics[width=0.30\textwidth]{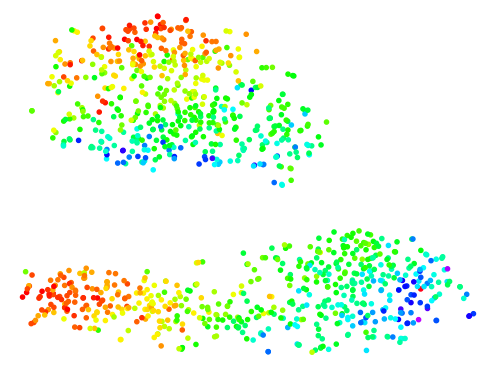}}
  \subfloat[NumCLIP (Ours)]{\includegraphics[width=0.30\textwidth]{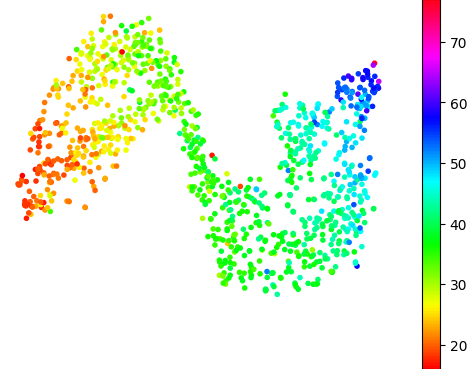}}
  \captionsetup{font=footnotesize}
  \caption{t-SNE visualisation of 512D image feature of CLIP on MORPH II}
  \label{fig:t-sne}
\end{figure}

Referring back to Table~\ref{table:ablation}, we can see that both the label conversion from pure numbers into common linguistic concepts, and fine-grained cross-modal ranking regularisation could further improve model performance and the joint combination of these modules makes the best performance. 
We provide t-SNE visualisation on the MORPH II dataset. Figure~\ref{fig:t-sne} shows that vanilla CLIP has an unordered feature representation among images with different ranks (ages), while NumCLIP could learn an ordinal and continuous representation that captures the intrinsic sample orders w.r.t the regression ranks (ages).
We also analyse the effect of different label distance functions on the model performance. As listed in Table~\ref{table:ablation on parameter}, the model performance is consistent with different choices and absolute label distance achieves the best performance.

  \begin{table*}[tb]
      \begin{minipage}[t]{0.505\textwidth}
    \centering\small
        \caption{Ablation study of NumCLIP on MORPH II under full-data setting.}
        \label{table:ablation}
          \renewcommand\tabcolsep{3pt}
        \scalebox{0.8}{
          \begin{tabular}{lcc|ccc}
            \toprule
            \multicolumn{2}{c}{Ablation Study}& {Baseline}$^*$ & (a)  & (b) & (c)  \\
            \midrule
            \multicolumn{2}{c}{Linguistic Concept} &\XSolidBrush  &\XSolidBrush &\Checkmark &\Checkmark   \\
            \multicolumn{2}{c}{Regularizer Loss} &\XSolidBrush  &\Checkmark &\XSolidBrush &\Checkmark  \\
            \midrule
            \multirow{1}{*}{} &MAE($\downarrow$)   &2.25 &2.15  &2.16 &\textbf{2.08}  \\
            \bottomrule
         \end{tabular}
         }
    \end{minipage}
  \hfill
 \begin{minipage}[t]{0.445\textwidth}
    \centering\small
        \caption{Ablation study of NumCLIP on selection of label distance functions.}
        \label{table:ablation on parameter}
          \renewcommand\tabcolsep{3pt}
         \scalebox{0.95}{
            \begin{tabular}{cc}
            \toprule
            Distance Function & MAE  \\
            $\lvert y_i - y_j \rvert$ & \textbf{2.08} \\
            $\sqrt{y_i - y_j}$ & \textbf{2.12} \\
            $(y_i - y_j)^{2}$ & \textbf{2.11}   \\
            \bottomrule
        \end{tabular}}
    \end{minipage}
  \end{table*}

% \vspace{-0.35cm}

\section{Discussions and Conclusions}
In this study, we have presented NumCLIP to teach CLIP with a strong number sense for ordinal regression, extending CLIP's potential in a new scenario. We first point out two major reasons for the limited performance of vanilla CLIP-based ordinal regression methods, namely the insufficient numerical training captions and ineffective training objective. To address these issues, we adopt a coarse-to-fine paradigm to reduce the learning difficulty on top of specially designed modules, which is achieved by performing intermediate classification first and then refining the prediction.
Common linguistic concepts converted from pure numbers are served as the intermediate labels for coarse classification to mitigate the label insufficiency issue. Fine-grained cross-modal ranking-based feature regularisation is proposed to subsequently refine the coarse linguistic labels with the guidance of ordinal property, through which both semantic and ordinal alignment are achieved in CLIP's feature space.
Extensive experimental results show that NumCLIP obtains competitive performance in general ordinal regression tasks, with 10\% overall accuracy improvement on historical image dating, 3.83\% overall accuracy improvement on image aesthetics assessment, and 1.33 MAE reduction on age estimation under 1-shot setting.

\section*{Acknowledgements}
This work is partially supported by a research grant
from NSFC under Grant 62306254, a research grant from the Beijing Institute of Collaborative Innovation (BICI) in collaboration with HKUST under Grant HCIC-004, and a project of Hetao Shenzhen-Hong Kong Science and Technology Innovation Cooperation Zone (HZQB-KCZYB-2020083).

% \clearpage  % TODO REVIEW/FINAL: This \clearpage needs to be removed from both review and camera-ready versions.

% ---- Bibliography ----
%
% BibTeX users should specify bibliography style 'splncs04'.
% References will then be sorted and formatted in the correct style.
%
\bibliographystyle{splncs04}
\bibliography{main}
\end{document}